\documentclass[twoside]{article}
\usepackage[a4paper]{geometry}
\usepackage[latin1]{inputenc} 
\usepackage[T1]{fontenc} 
\usepackage{hyperref}
\usepackage{graphicx, amsmath, theorem, amssymb, algorithmic}

\begin{document}

\newcommand{\proof}{{\bf Proof:~}}
\newcommand{\uio}{$\Sigma^{(0)}$}
\newcommand{\uioo}{$\Sigma^{(1)}$}
\newcommand{\uiok}{$\Sigma^{(k)}$}
\newcommand{\eorc}{$EORC$}
\newcommand{\uiokm}{$\Sigma^{(\overline{k})}$}
\newcommand{\uioK}{$\Sigma^{(k+1)}$}

\title{Complete analytic solution to Brownian unicycle dynamics}

\author{Agostino Martinelli}

\maketitle

\begin{abstract}
This paper derives a complete analytical solution for the probability distribution of the configuration of a non-holonomic vehicle that moves in two spatial dimensions by satisfying the unicycle kinematic constraints and in presence of Brownian noises. In contrast to previous solutions, the one here derived holds even in the case of arbitrary linear and angular speed. This solution is obtained by deriving the analytical expression of any-order moment of the probability distribution. To the best of our knowledge, an analytical expression for any-order moment that holds even in the case of arbitrary linear and angular speed, has never been derived before. To compute these moments, a direct integration of the Langevin equation is carried out and each moment is expressed as a multiple integral of the deterministic motion (i.e., the known motion that would result in absence of noise). 
For the special case when the ratio between the linear and angular speed is constant, the multiple integrals can be easily solved  and expressed as the real or the imaginary part of suitable analytic functions. As an application of the derived analytical results, the paper investigates the diffusivity of the considered Brownian motion for constant and for arbitrary time-dependent linear and angular speed.\end{abstract}
{\bf Keywords:} Brownian motion, Stochastic dynamics, Fokker-Plank equation without detailed balance, Stochastic processes, Localization.

\newpage
\tableofcontents

\newpage

\section{Introduction}

In recent years a great interest has been devoted to investigate the statistical properties of the motion of active (or self-propelled) particles. These particles differ from passive particles since they move under the action of an internal force. Typical examples of these particles are Brownian motors \cite{Fre13,Han09,Rei02}, biological and artificial microswimmers \cite{Dre05,Lau09,Leo09,Sok07,Tie08,Tie08b,Tro08}, macroscopic animals \cite{Baz10,Cou02,Niw94} and even pedestrians \cite{Tur12}.

In the last decade, this investigation has also been addressed in the framework of mobile robotics, in particular by analyzing the $2D$ cases of the unicyle, cart, and car \cite{Zho03,Zho04}, and the $3D$ case of flexible needle steering \cite{Park05,Park10}. In these works, analytical solutions are obtained by directly solving the corresponding Fokker-Planck equations by using the Fourier transform. Another method proposed for obtaining a closed-form solution is the use of exponential coordinates \cite{Long12,Park08}. In all these works, the benefit of having closed-form solutions is clearly illustrated by introducing motion planning methods based on them.

In this paper we want to deal with wheeled robots, which represent a special case of active particles moving in a two dimensional space. Their configuration is characterized by a 3D-vector (2 components characterize their position and one their orientation). On the other hand, in most of cases, a wheeled robot cannot move along any direction since it must satisfy the so-called non-holonomic constraints \cite{Lat91}. Our stochastic motion model (presented in section \ref{SectionStochasticMotionModel}) differs from the one considered in \cite{Long12,Park08,Zho03,Zho04} since we assume independent errors acting on the linear and angular components. The model adopted in \cite{Long12,Park08,Zho03,Zho04} refers to the case of a differential drive system where the errors acting on each wheel are independent.

Starting from our stochastic motion model, we compute the analytical expression of any-order moment of the probability distribution (section \ref{SectionComputationPdf}). Specifically, each moment is expressed as a multiple integral of the deterministic motion performed by the mobile robot (i.e., the known motion that would result in absence of noise). In other words, the expression is a multiple integral on two time-dependent functions, which describe the time behaviour of the deterministic linear and angular speed. This allows obtaining any order moment for any deterministic motion, i.e., when the mobile robot is propelled by arbitrary time-dependent linear and angular speed. For the special case when the ratio between these two speeds is constant, the multiple integrals can be expressed as the real or the imaginary part of suitable analytic functions. In section \ref{SectionDiffusivity} we show the power of the derived analytical results by investigating the diffusivity of the considered Brownian motion for constant and for arbitrary time-dependent linear and angular speed.
Conclusions are provided in section \ref{SectionConclusion}.

The interest in deriving the statistical properties of the motion of a wheeled robot comes from the possibility of improving its localization (both in precision and speed). The localization is a fundamental problem in mobile robotics which must be solved to autonomously and safely navigate. It is a non linear estimation problem. In most of cases, the localization is carried out by using nonlinear filters (e.g., Extended and Unscented Kalman Filters, Particle filters, etc). On the other hand, non linear filters are strictly connected with the solution of partial differential equations (see \cite{Cha00,Dau86,Dau87,Dau05} and the next section \ref{SectionStochasticMotionModel}). Very easily speaking, for a dynamic stochastic process described by stochastic differential equations, it is possible to introduce a probability distribution which provides the probability that the process takes a given set of values. This probability distribution must satisfy a partial differential equation. In this sense, a non linear filter could be considered as a numerical solution of this partial differential equation. In the specific case of localization, the dynamic process is the motion of the robot together with the noisy data delivered by its sensors.

We already computed the statistics up to the second order for a similar motion model \cite{TRA02,ICRA02} (in the expressions in \cite{TRA02} there are some typos, corrected in \cite{ICRA02}). Here, we extend the computation in \cite{TRA02} by including any-order statistics. We also show that the expressions hold for a much more general model (the one described in section \ref{SectionStochasticMotionModel}) and not only for a simple specific odometry error model (as shown in \cite{TRA02}). As a result, this paper provides a complete analytical solution to a Fokker-Planck equation for which the detailed balance condition is not verified\footnote{Note that, deriving any-order moment, corresponds to analytically derive the probability density distribution \cite{Papoulis}.}.

Note that the goal of this paper is to analytically compute the moments of the probability distribution up to any order for the configuration of a non-holonomic vehicle. As it will be seen, this is a very hard task from a computational point of view. These analytical results could be very useful to improve the localization which is, as previously mentioned, a numerical solution of a partial differential equation.

\section{Stochastic motion model}\label{SectionStochasticMotionModel}

We consider a mobile robot which moves in a $2D-$environment. The configuration of the mobile robot is characterized by its position and orientation. In Cartesian coordinates, we denote them by $\mbox{\boldmath $r$}\equiv [x, ~y]^T$ and $\theta$, respectively. We assume that the motion of the mobile robot satisfies the {\it unicycle} model \cite{Lat91}, which is the most general nonholonomic model for $2D-$motions, where the shift of the mobile robot can only occur along one direction ($[\cos\theta, ~\sin\theta]^T$):

\begin{equation}\label{EquationUnicycle}
\left[
\begin{array}{cc}
\dot{x}=&v\cos\theta\\
\dot{y}=&v\sin\theta\\
\dot{\theta}=&\omega
\end{array}
\right.
\end{equation}

\noindent The quantities $v=v(t)$ and $\omega=\omega(t)$ are the linear and angular speed, respectively. We assume that these functions are known and we call the motion that results from them the {\it deterministic} motion. Now we want to introduce a stochastic model that generalizes equation (\ref{EquationUnicycle}) by accounting a Gaussian white noise. We assume that also the noise satisfies the same nonholonomic constraint. The stochastic differential equations are:

\begin{equation}\label{EquationUnicycleStochastic}
\left[
\begin{array}{cc}
dx(t)=&\cos\theta(t)~[v(t) dt + \alpha(t) dw^r(t)]\\
dy(t)=&\sin\theta(t)~[v(t) dt + \alpha(t) dw^r(t)]\\
d\theta(t)=&\omega(t) dt + \beta(t) dw^{\theta}(t)
\end{array}
\right.
\end{equation}

\noindent where $[dw^r(t),~dw^{\theta}(t)]^T$ is a standard Wiener process of dimension two \cite{Oks03}. The functions $\alpha(t)$ and $\beta(t)$ are modeled according to the following physical requirement (diffusion in the overdamped regime). We require that, when the mobile robot moves during the infinitesimal time interval $dt$, the shift is a random Gaussian variable, whose variance increases linearly with the traveled distance. This is obtained by setting $\alpha(t)=\sqrt{K_r |v(t)|}$, with $K_r$ a positive parameter which characterizes our system (interaction robot-environment). Similarly, we require that the rotation accomplished during the same time interval $dt$ is a random Gaussian variable, whose variance increases linearly with the traveled distance. Hence, we set  $\beta(t)=\sqrt{K_{\theta} |v(t)|}$, with $K_{\theta}$ another positive parameter which characterizes our system (interaction robot-environment).
The equations in (\ref{EquationUnicycleStochastic}) are the Langevin equations in the overdamped limit. From now on, we assume that the mobile robot can only move ahead\footnote{Note that this constraint does not limit the robot motion: for instance, a {\it go and back} motion is easily obtained by accomplishing $180deg$ rotations}, i.e., $v(t)\ge 0$. 
We remind the reader that $v(t)$ is a deterministic and known function of time. The curve length $ds=v(t) dt$ is the curve length of the deterministic motion, which is independent of the Wiener process $[dw^r(t),~dw^{\theta}(t)]^T$. Hence, we are allowed to use the deterministic curve length $ds=v(t) dt$ instead of the time $t$. By denoting the ratio between the angular and the linear speed by $\mu$, equations in (\ref{EquationUnicycleStochastic}) read:

\begin{equation}\label{EquationUnicycleStochasticS}
\left[
\begin{array}{cc}
dx(s)=&\cos\theta(s)~[ds + K_r^{\frac{1}{2}} dw^r(s)]\\
dy(s)=&\sin\theta(s)~[ds + K_r^{\frac{1}{2}} dw^r(s)]\\
d\theta(s)=&\mu(s) ds + K_{\theta}^{\frac{1}{2}} dw^{\theta}(s)
\end{array}
\right.
\end{equation}

\noindent The associated Smoluchowski equation is:

\begin{equation}\label{EquationSmoluchowski}
\frac{\partial p}{\partial s} = - \mbox{\boldmath $\nabla$} \cdot (\mbox{\boldmath $\mathcal{D}_1$} p) + \mbox{\boldmath $\nabla$} \cdot (\mbox{\boldmath $\mathcal{D}_2$} \mbox{\boldmath $\nabla$}p)=0
\end{equation}

\noindent where:

\begin{itemize}

\item $p=p(x,~y, ~\theta; ~s(t))$ is the probability density for the mobile robot at the configuration $(x,~y, ~\theta)$ and at time $t$;

\item $\mbox{\boldmath $\mathcal{D}_1$}=[\cos\theta, ~\sin\theta, ~\mu]^T$ is the drift vector;

\item $\mbox{\boldmath $\mathcal{D}_2$}=\frac{1}{2}\left[
\begin{array}{ccc}
K_r \cos^2\theta& K_r \cos\theta\sin\theta& 0\\
K_r \cos\theta\sin\theta&K_r \sin^2\theta&  0\\
0&0&K_{\theta}\end{array}
\right]
$ is the diffusion tensor.

\end{itemize}

\noindent Our goal is to obtain the probability density $p(x,~y, ~\theta; ~s)$. This will be done in the next two sections.

%
%

%
%

\section{Computation of the probability distribution}\label{SectionComputationPdf}


The probability density $p(x,~y, ~\theta;~ s)$ satisfies the Smoluchowski partial differential equation in (\ref{EquationSmoluchowski}), which is a special case of the Fokker-Planck equation. Since the detailed balance condition is not satisfied \cite{Ris89}, we follow a different procedure to compute $p(x,~y, ~\theta; ~s)$. Specifically, we use the Langevin equation in (\ref{EquationUnicycleStochasticS}) to compute the moments, up to any order, of the probability distribution. First of all, we remark that the third equation in (\ref{EquationUnicycleStochasticS}) is independent of the first two, is independent of $\theta$ and is linear in $dw^{\theta}$. As a result, the probability density only in terms of $\theta$, i.e., the probability $P_{\theta}(\theta; ~s) \equiv \int dx \int dy~p(x,~y, ~\theta; ~s)$, is a Gaussian distribution with mean value $\theta_0+\int_0^s ds' \mu(s')$ and variance $K_{\theta} s$, where $\theta_0$ is the initial orientation. This same result could also be obtained by integrating (\ref{EquationSmoluchowski}) in $x$ and $y$ and by using the divergence theorem. In other words, we have the following distribution for the orientation at a given value of $s$:

\begin{equation}\label{EquationTheta(s)}
\theta(s)=\mathcal{N}(\overline{\theta}(s), ~K_{\theta}s)
\end{equation}

\noindent where $\mathcal{N}(\cdot, ~\cdot)$ denotes the normal distribution with mean value and variance the first and the second argument; $\overline{\theta}(s) \equiv \theta_0+\int_0^s ds' ~\mu(s')$. Note that, according to the unicycle model, a trajectory is completely characterized by its starting point and by the orientation vs the curve length, i.e., by the function $\theta(s)$. In the following, we will call the deterministic function $\overline{\theta}(s)$, the {\it deterministic} trajectory.

\subsection{First and second-order statistics}\label{SubSectionSecondOrder}

Let us consider the first equation in (\ref{EquationUnicycleStochasticS}). We compute the expression of $x(s)$ by a direct integration. We divide the interval $(0, ~s)$ in $N$ equal segments, $\delta s\equiv \frac{s}{N}$. We have:

\begin{equation}\label{EquationXLim}
x(s)=\lim_{N\rightarrow\infty} \sum_{j=1}^N (\delta s + \epsilon_j) \cos\theta_j
\end{equation}

\noindent where $\epsilon_j$ is a random Gaussian variable satisfying $\left<\epsilon_j\right>=0$, $\left<\epsilon_j ~\epsilon_k\right>=\delta_{jk} K_r \delta s$ for $j,k = 1,\cdots, N$ ($\delta_{jk}$ is the Kronecker delta) and $\theta_j \equiv \theta(j \delta s)$. On the other hand, we have:

\begin{equation}\label{EquationThetaLim}
\theta_j= \overline{\theta}(j \delta s)+\sum_{m=1}^j \delta \theta_m \equiv \overline{\theta}_j + \Delta\theta_j
\end{equation}

\noindent where, according to our stochastic model in (\ref{EquationUnicycleStochasticS}), $\delta\theta_j$ is a random Gaussian variable satisfying $\left<\delta\theta_j\right>=0$, $\left<\delta\theta_j ~\epsilon_k\right>=0$ and $\left<\delta\theta_j ~\delta\theta_k\right>=\delta_{jk} K_{\theta} \delta s$ for $j,k = 1,\cdots, N$. As a result, $\Delta\theta_j\equiv \sum_{m=1}^j \delta \theta_m$ is also a random Gaussian variable satisfying $\left<\Delta\theta_j\right>=0$, $\left<\Delta\theta_j ~\Delta\theta_k\right>=\left[
\begin{array}{cc}
K_{\theta} j \delta s & if ~j\le k\\
K_{\theta} k \delta s & if ~j>k\\
\end{array}
\right.
$. Starting from (\ref{EquationXLim}) and (\ref{EquationThetaLim}) and by remarking that $\left<\cos\Delta\theta_j\right>=e^{-\frac{k_{\theta}j\delta s}{2}}$ and $\left<\sin\Delta\theta_j\right>=0$, it is easy to obtain the mean value of $x(s)$. We have:

\begin{equation}\label{EquationXMean}
\left<x(s)\right>=\lim_{N\rightarrow\infty} \sum_{j=1}^N (\delta s + \left<\epsilon_j\right>) \left<\cos\theta_j\right>=
\end{equation}
\[=\lim_{N\rightarrow\infty} \sum_{j=1}^N \delta s ~\cos\overline{\theta}_j~e^{-\frac{k_{\theta}j\delta s}{2}}=\int_0^s ds'~\cos\overline{\theta}(s')~e^{-\frac{K_{\theta} s'}{2}}\]

\noindent Similarly, it is possible to obtain the mean value of $y(s)$:

\begin{equation}\label{EquationYMean}
\left<y(s)\right>=\int_0^s ds'~\sin\overline{\theta}(s')~e^{-\frac{K_{\theta} s'}{2}}
\end{equation}

\noindent Finally, in a similar manner, but with some more computation, we obtain all the second-order moments:

\begin{equation}\label{EquationX^2}
\left<x(s)^2\right>=\int_0^sds'
\int_0^{s-s'}ds''
e^{-\frac{K_{\theta}s''}{2}}
\left\{
(1+\chi_c(s'))
\right.
\end{equation}
\[
\left.
\cos[\overline{\theta}(s'+s'')-\overline{\theta}(s')]-\chi_s(s')\sin[\overline{\theta}(s'+s'')-\overline{\theta}(s')]
\right\} +
\]
\[
+\frac{K_r}{2}\left[ s+\int_0^sds'~\chi_c(s')\right]
\]

\begin{equation}\label{EquationY^2}
\left<y(s)^2\right>=\int_0^sds'
\int_0^{s-s'}ds''
e^{-\frac{K_{\theta}s''}{2}}
\left\{
(1-\chi_c(s'))
\right.
\end{equation}
\[
\left.
\cos[\overline{\theta}(s'+s'')-\overline{\theta}(s')]+\chi_s(s')\sin[\overline{\theta}(s'+s'')-\overline{\theta}(s')]
\right\} +
\]
\[
+\frac{K_r}{2}\left[ s-\int_0^sds'~\chi_c(s')\right]
\]

\begin{equation}\label{EquationXY}
\left<x(s)~y(s)\right>=\int_0^sds'
\int_0^{s-s'}ds''
e^{-\frac{K_{\theta}s''}{2}}
\left\{
\chi_s(s')
\right.
\end{equation}
\[
\left.
\cos[\overline{\theta}(s'+s'')-\overline{\theta}(s')]+\chi_c(s')\sin[\overline{\theta}(s'+s'')-\overline{\theta}(s')]
\right\} +
\]
\[
+\frac{K_r}{2}\int_0^sds'~\chi_s(s')
\]

\begin{equation}\label{EquationXTheta}
\sigma_{x\theta}(s) \equiv \left<x(s)~\theta(s)\right>-\left<x(s)\right>\overline{\theta}(s)=2K_{\theta}\frac{\partial \left<y(s)\right>}{\partial K_{\theta}}
\end{equation}

\begin{equation}\label{EquationYTheta}
\sigma_{y\theta}(s) \equiv \left<y(s)~\theta(s)\right>-\left<y(s)\right>\overline{\theta}(s)=-2K_{\theta}\frac{\partial \left<x(s)\right>}{\partial K_{\theta}}
\end{equation}

\noindent where $\chi_c(s') \equiv \cos[2\overline{\theta}(s')]~e^{-2K_{\theta}s'}$ and
$\chi_s(s') \equiv \sin[2\overline{\theta}(s')]~e^{-2K_{\theta}s'}$.

Obtaining the expression of higher-order moments will demand more tricky computation and will be dealt separately, in the next subsection. Here we conclude by considering the quantity $D(s)^2 \equiv x(s)^2 + y(s)^2$, which provides the time-evolution of the square of the distance of the mobile robot from its initial position. From (\ref{EquationX^2}) and (\ref{EquationY^2}) we obtain a simple expression for its mean value:

\begin{equation}\label{EquationD^2}
\left<D(s)^2\right>=
\end{equation}
\[
K_r s+2\int_0^sds'
\int_0^{s-s'}ds''
e^{-\frac{K_{\theta}s''}{2}}
\cos[\overline{\theta}(s'+s'')-\overline{\theta}(s')]
\]

\subsection{Computation of any-order moment}\label{SubSectionSecondOrder}

We introduce the following two complex random quantities:

\begin{equation}\label{EquationUV}
u(s)\equiv \lim_{N\rightarrow\infty} \sum_{j=1}^N (\delta s + \epsilon_j) e^{i\theta_j};~
w(s)\equiv \lim_{N\rightarrow\infty} \sum_{j=1}^N (\delta s + \epsilon_j) e^{-i\theta_j}
\end{equation}

\noindent From (\ref{EquationXLim}) and (\ref{EquationUV}) it is immediate to realize that $x(s)=\frac{u(s)+w(s)}{2}$. Similarly we have $y(s)=\frac{u(s)-w(s)}{2i}$. Hence, in order to compute any-order moment which involves $x(s)$ and $y(s)$ it  suffices to compute $\left<u(s)^p ~w(s)^q\right>$ for any $p,q\in \mathcal{N}$. The computation of this quantity requires several tricky steps, which are provided in appendix \ref{AppendixUV}. The key is to separate all the independent random quantities in order to compute their mean values. This is obtained by arranging all the sums in a suitable manner (see appendix \ref{AppendixUV} for all the details). We have:

\begin{equation}\label{Equation<UV>Computed}
\left<u(s)^p ~w(s)^q\right>=\sum_{n=0}^{\lfloor \frac{p+q}{2} \rfloor} K_r^n \sum_{l=0}^{2n} {{p}\choose{2n-l}} {{q}\choose{l}}
\end{equation}
\[
\sum_{m} {{2n-l}\choose{m}} {{l}\choose{m}} m! (2n-l-m-1)!! (l-m-1)!! ~s^m
\]
\[
e^{i(p-q)\theta_0}\left(\frac{2n-l-m}{2}\right)!\left(\frac{l-m}{2}\right)! (p-2n+l)!  (q-l)!
\]
\[
\sum_{\mbox{\boldmath $c$}}
\int_0^s ds_1 \int_{s_1}^s ds_2\cdots \int_{s_{\beta-1}}^s ds_{\beta}
exp\left\{ \sum_{b=0}^{\beta-1}\left[ i(p-q + \Phi_b) \cdot \right. \right.
\]
\[
 \left. \left. \cdot [\overline{\theta}(s_{b+1})-\overline{\theta}(s_b)]-\frac{(p-q + \Phi_b)^2 (s_{b+1}-s_b)K_{\theta}}{2}\right]\right\}
\]
\noindent where:

\begin{itemize}

\item $\lfloor \frac{p+q}{2} \rfloor$ is the largest integer not greater than $\frac{p+q}{2}$;

\item the second sum on $l$ (i.e., $\sum_{l=0}^{2n}$) is restricted to the values of $l$ for which $2n-l\le p$ and $l\le q$ (or, equivalently, we are using the convention that ${{x}\choose{y}}=0$ when $y>x$);

\item the sum on $m$ (i.e., $\sum_m$) goes from $0$ to the minimum between $l$ and $2n-l$ and it is restricted to the integers $m$ with the same parity of $l$ (hence, both $\frac{2n-l-m}{2}$ and $\frac{l-m}{2}$ are integers);

\item the symbols "!" and "!!" denote the factorial and the double factorial, respectively \cite{Cal09} (note that $0!=0!!=(-1)!!=1$);

\item $\beta=p+q-n-m$ is the dimension of the remaining multiple integral (note that a multiple integral of dimension $m$ has already been computed and provided the term $s^m$);

\item the sum on $\mbox{\boldmath $c$}$, i.e., $\sum_{\mbox{\boldmath $c$}}$, is the sum over all the vectors $\mbox{\boldmath $c$}$ of dimension $\beta$, whose entries are $-2$, $-1$, $1$ and $2$: specifically, each vector $\mbox{\boldmath $c$}$ has $\frac{l-m}{2}$ entries equal to $2$, $q-l$ equal to $1$, $p-2n+l$ equal to $-1$ and $\frac{2n-l-m}{2}$ equal to $-2$ (note that the sum $\sum_{\mbox{\boldmath $c$}}$ consists of  ${{\beta}\choose{q-l}} {{\beta-q+l}\choose{p-2n+l}} {{n-m}\choose{\frac{l-m}{2}}}$ addends);

\item $\Phi_b \equiv \sum_{a=1}^b c_a$ (note that $\Phi_{\beta}=q-p$);

\item $s_0 \equiv 0$ and $\overline{\theta}(s_0) = \theta_0$ (note that $s_0$ is not a variable of integration).

\end{itemize}

\noindent In order to complete the derivation of the statistics for our problem, we need to compute any-order moment which also involves the orientation $\theta$. We provide the formula for the quantity $\left<u(s)^p ~w(s)^q~\tilde{\theta}(s)^r\right>$, where $\tilde{\theta}(s) \equiv \theta(s)-\overline{\theta}(s)$. The details of this computation are provided in appendix \ref{AppendixUVTheta}.  We have, for any $p,q,r\in \mathcal{N}$:

\begin{equation}\label{Equation<UVT>Computed}
\left<u(s)^p ~w(s)^q~\tilde{\theta}(s)^r\right>=\sum_{n=0}^{\lfloor \frac{p+q}{2} \rfloor} K_r^n \sum_{l=0}^{2n} {{p}\choose{2n-l}} {{q}\choose{l}}
\end{equation}
\[
\sum_{m} {{2n-l}\choose{m}} {{l}\choose{m}} m! (2n-l-m-1)!! (l-m-1)!! ~s^m ~e^{i(p-q)\theta_0}
\]
\[
\left(\frac{2n-l-m}{2}\right)!\left(\frac{l-m}{2}\right)! (p-2n+l)!  (q-l)! \sum_{\mbox{\boldmath $c$}} \sum_{\mbox{\boldmath $\gamma$}}
\]
\[
\frac{r! (\gamma_{\beta+1}-1)!! K_{\theta}^{\frac{r}{2}}}{\prod_{b=0}^{\beta}\gamma_{b+1}!}\int_0^s ds_1 \int_{s_1}^s ds_2 \cdots \int_{s_{\beta-1}}^s ds_{\beta} (s-s_{\beta})^{\frac{\gamma_{\beta+1}}{2}}
\]
\[
exp\left\{ i\sum_{b=0}^{\beta-1}(p-q + \Phi_b) [\overline{\theta}(s_{b+1})-\overline{\theta}(s_b)]\right\} \cdot
\]
\[
\cdot \prod_{b=0}^{\beta-1}
\left\{ (s_{b+1}-s_b)^{\frac{\gamma_{b+1}}{2}}exp\left[-\frac{(p-q + \Phi_b)^2 K_{\theta} (s_{b+1}-s_b)}{2} \right] \right.
\]
\[
\left. \sum_{a=0}^{\lfloor\frac{\gamma_{b+1}}{2}\rfloor}\frac{\gamma_{b+1}!\left(i (p-q + \Phi_b) \sqrt{K_{\theta} (s_{b+1}-s_b)}\right)^{\gamma_{b+1}-2a}}{a!(\gamma_{b+1}-2a)!2^a}\right\}
\]

\noindent where the sum over $\mbox{\boldmath $\gamma$}$, i.e. $\sum_{\mbox{\boldmath $\gamma$}}$, is the sum over all the vectors $\mbox{\boldmath $\gamma$}=[\gamma_1,\gamma_2,\cdots,\gamma_{\beta},\gamma_{\beta+1}]$, where $\gamma_b$ ($b=1,\cdots,\beta$) are positive integers and $\gamma_{\beta+1}$ is a positive integer with even parity. Additionally, they satisfy the constraint: $\sum_{b=1}^{\beta+1}\gamma_b=r$.

\section{Diffusivity for constant and for arbitrary time-dependent linear and angular speed}\label{SectionDiffusivity}

In this section we want to illustrate the power of the formulas derived in the previous section, which hold for any time-dependent linear and angular speed, to compute the statistics in several specific cases. In particular, we compare the results obtained by using our formulas with the results that it is possible to obtain by running Monte Carlo simulations. We generate many trials of random motion via Monte Carlo simulations according to the motion model in (\ref{EquationUnicycleStochasticS}). The initial configuration is the same for all the trials and it is $x(0)=y(0)=\theta(0)=0$. The final curve length is also the same and it is $s=1$. We consider several values for the two parameters $K_{\theta}$ and $K_r$ and we set $\mu(s)$ constant in \ref{SubSectionConstTorque} and variable in \ref{SubSectionArbitraryTorque}. We will see that, as the number of trials increases, the statistical properties of the motion directly obtained from the trials approach the ones obtained by using our formulas. In particular, we will refer to the motion diffusivity and we compute the moments up to the forth order. We start by computing for each trial, the square of the final distance from the origin, i.e. the distance at $s=1$:

\begin{equation}
D(1)^2=x(1)^2+y(1)^2 
\end{equation}

\noindent Once we have the previous quantity for many trials, we compute its mean value and its variance. On the other hand, our formulas allow us to directly obtain both the mean value and the variance. Specifically, equation (\ref{EquationD^2}) provides the mean value. Regarding the variance we have: $\sigma_{D(s)^2}\equiv \left< D(s)^4\right>-\left< D(s)^2\right>^2$. Additionally, $D(s)^4=x(s)^4+2x(s)^2y(s)^2+y(s)^4=u(s)^2w(s)^2$, where we used  $x(s)=\frac{u(s)+w(s)}{2}$ and $y(s)=\frac{u(s)-w(s)}{2i}$. The quantity $\left< u(s)^2w(s)^2\right>$ can be easily computed by using equation (\ref{Equation<UV>Computed}) with $p=q=2$ (see appendix \ref{AppendixD4} for the details). Its analytical expression includes the following six terms: $n=0$, $l=0$, $m=0$ ($C_{000}$); $n=1$, $l=0$, $m=0$ ($C_{100}$); $n=1$, $l=1$, $m=1$ ($C_{111}$); $n=1$, $l=2$, $m=0$ ($C_{120}$); $n=2$, $l=2$, $m=0$ ($C_{220}$) and $n=2$, $l=2$, $m=2$ ($C_{222}$). We have:

\begin{equation}\label{EquationD^4}
\left< D(s)^4 \right>=C_{000}+C_{100}+C_{111}+C_{120}+C_{220}+C_{222}
\end{equation}

\noindent See appendix \ref{AppendixD4} for the analytical expression of the previous terms. In the next two subsections, we compare the results for the mean value and the variance of $D(1)^2$ obtained by running many trials and by using our analytic expressions. In \ref{SubSectionConstTorque} we refer to a constant ratio between the linear and the angular speed while in \ref{SubSectionArbitraryTorque} we refer to a generic varying ratio. Note that this analysis is a test for the $4^{th}$ order statistics of the considered Brownian motion.

\subsection{Constant ratio between linear and angular speed}\label{SubSectionConstTorque}

When $\mu(s)=\mu_0$ we have that $\overline{\theta}(s)$ is linear in $s$:

\begin{equation}
\overline{\theta}(s)=\theta_0+\mu_0s
\end{equation}

\noindent and, the corresponding active trajectory, is a circumference with radius $\frac{1}{\mu_0}$ (note that when the angular speed vanishes, the radius becomes infinite and the active trajectory becomes a straight line). The computation of the integrals in (\ref{Equation<UV>Computed}) and (\ref{Equation<UVT>Computed}) is immediate.
As a result, the expression of $\left<D(s)^2\right>$ can be easily provided in terms of $\mu_0$, $s$, $K_r$ and $K_{\theta}$. Let us introduce the following complex quantity:

\begin{equation}
z \equiv -\frac{K_{\theta}}{2}+i\mu_0
\end{equation}

\noindent By a direct analytical computation of the double integral in (\ref{EquationD^2}) we easily obtain:

\begin{equation}\label{EquationD^2Circ}
\left<D(s)^2\right>= K_r s+2\Re \left\{\frac{ e^{zs}-1-zs}{z^2}\right\}
\end{equation}

\noindent  where the symbol $\Re\{\cdot\}$ denotes the real part of a given complex quantity.

We do not provide here the analytical expression for $\left< D(s)^4 \right>$. We only remark that all the integrals appearing in the expressions in appendix \ref{AppendixD4} can be computed in a similar manner and are expressed as the real part of suitable analytic functions of the following complex quantities: $z$, $-\frac{3K_{\theta}}{2}+i\mu_0$ and $-2K_{\theta}+ 2i\mu_0$.

\begin{figure}[htbp]
\begin{center}
\includegraphics[width=.8\columnwidth]{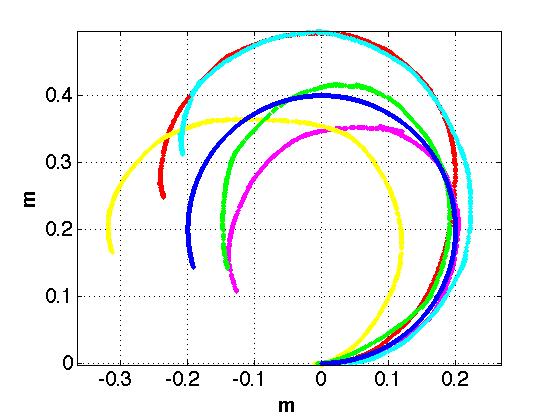}
\caption{Trajectories generated in the case of constant ratio between linear and angular speed. The blue line is the trajectory generated without noise (i.e, $K_{\theta}=K_r=0$) and the other lines are five trajectories obtained by setting $K_{\theta}=K_r=0.01m$.} \label{FigTrajCirc}
\end{center}
\end{figure}

In the following, we illustrate some numerical results obtained by setting $\mu(s)=\mu_0=5$. We consider two cases of Brownian noises, the former is characterized by $K_{\theta}=K_r=0.01m$ and the latter by $K_{\theta}=K_r=1m$. Figure \ref{FigTrajCirc} displays the trajectory generated without noise (blue line) together with five trajectories obtained by setting $K_{\theta}=K_r=0.01m$.
For the trajectory without noise, the final $D(1)^2$ is equal to $0.0573~m^2$. We use the expression in (\ref{EquationD^2Circ}) to compute the mean value and a similar expression, which depends on the three mentioned complex quantities $z$, $-\frac{3K_{\theta}}{2}+i\mu_0$ and $-2K_{\theta}+ 2i\mu_0$, to compute the variance. We obtain the following values: $\left< D(1)^2 \right>=0.0680~m^2$ and $\sigma_{D(1)^2}^2=0.0012~m^4$ when 
$K_{\theta}=K_r=0.01m$ and $\left< D(1)^2 \right>=1.1130~m^2$ and $\sigma_{D(1)^2}^2=1.3052~m^4$ when 
$K_{\theta}=K_r=1m$.

\begin{figure}[htbp]
\begin{center}
\begin{tabular}{cc}
\includegraphics[width=.5 \columnwidth]{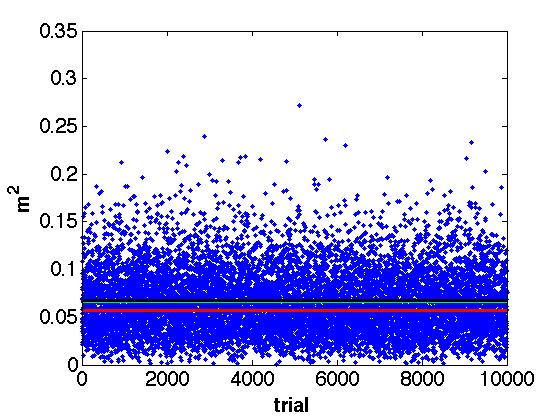}&
\includegraphics[width=.5 \columnwidth]{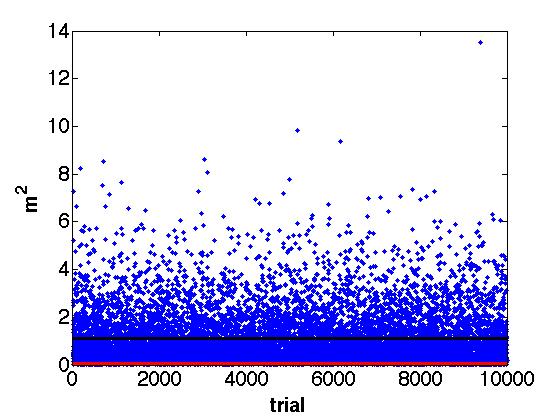}\\
\end{tabular}
\caption{Constant ratio between linear and angular speed. Values of $D(1)^2$ for $10^4$ trials together with the value of $D(1)^2$ for the trajectory without noise (red line), the value of $\left< D(1)^2 \right>$ obtained by averaging $D(1)^2$ on the trials (green line) and the value of $\left< D(1)^2 \right>$ obtained by our formulas (black line). $K_{\theta}=K_r=0.01m$ and $K_{\theta}=K_r=1m$ in the left and right image, respectively.}\label{FigStatCirc}
\end{center}
\end{figure}

We ran $10^5$ trials and we computed from them the mean value and the variance. Figure \ref{FigStatCirc} displays the values of $D(1)^2$ obtained in the first $10^4$ trials together with the value of $D(1)^2$ for the trajectory without noise (red line), the value of $\left< D(1)^2 \right>$ obtained by averaging $D(1)^2$ on the $10^4$ trials (green line) and the value of $\left< D(1)^2 \right>$ obtained by our formulas (black line). The image on the left refers to the case $K_{\theta}=K_r=0.01m$ and the image on the right to the case $K_{\theta}=K_r=1m$.

\noindent Table \ref{TableCirc} reports the values of $\left< D(1)^2 \right>$ and $\sigma_{D(1)^2}^2$ obtained by running $10^3$, $10^4$ and $10^5$ trials. It is possible to see that, as the number of trials increases, the results converge to the values obtained by using our formulas.

\begin{table}
\begin{center}
\begin{tabular}{|c|c|c|c|}
  \hline
  $K_{\theta}=K_r~ (m)$  &  trials & $\left< D(1)^2 \right>~(m^2)$ & $\sigma_{D(1)^2}^2~(m^4)$ \\  
   \hline
   \hline
   $0.01$ & $10^3$ & $0.0688$ & $0.0011$\\
   \hline
   $0.01$ & $10^4$ & $0.0664$ & $0.0012$\\
   \hline
   $0.01$ & $10^5$ & $0.0679$ & $0.0012$\\
   \hline
   \hline
   $1$ & $10^3$ & $1.1189$ & $1.3736$\\
   \hline
    $1$ & $10^4$ & $1.1124$ & $1.2767$\\
   \hline
    $1$ & $10^5$ & $1.1126$ & $1.3035$\\
   \hline
   \hline
\end{tabular}
\end{center}
\caption{Constant ratio between linear and angular speed. Values of $\left< D(1)^2 \right>$ and $\sigma_{D(1)^2}^2$ obtained by running $10^3$, $10^4$ and $10^5$ trials.}
\label{TableCirc}
\end{table}

\subsection{Arbitrary time-dependent ratio between linear and angular speed}\label{SubSectionArbitraryTorque}

\begin{figure}[htbp]
\begin{center}
\includegraphics[width=.8\columnwidth]{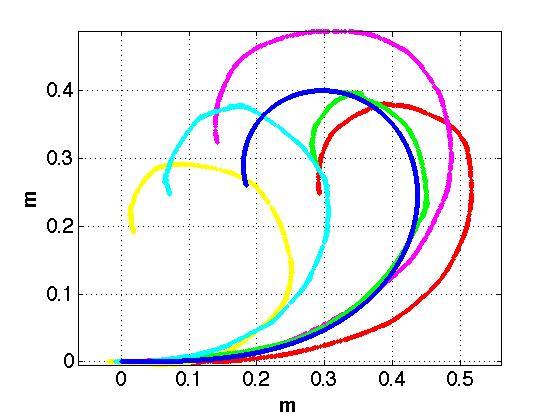}
\caption{Trajectories generated in the case of varying ratio between linear and angular speed. The blue line is the trajectory generated without noise (i.e, $K_{\theta}=K_r=0$) and the other lines are five trajectories obtained by setting $K_{\theta}=K_r=0.01m$.} \label{FigTraj}
\end{center}
\end{figure}

We consider now a generic function $\mu(s)$. Specifically, we consider several kinds of dependence on $s$ obtaining very similar results. In the following, we illustrate the results obtained by setting $\mu(s)=10~s$. As in the previous case, we consider the two cases of Brownian noises, i.e., characterized by $K_{\theta}=K_r=0.01m$ and by $K_{\theta}=K_r=1m$. Figure \ref{FigTraj} displays the trajectory generated without noise (blue line) together with five trajectories obtained by setting $K_{\theta}=K_r=0.01m$.
For the trajectory without noise, the final $D(1)^2$ is equal to $0.1021~m^2$. In this case we directly use the expressions in (\ref{EquationD^2}) in (\ref{EquationD^4}) and in appendix \ref{AppendixD4} by numerically computing all the integrals. We obtain the following values: $\left< D(1)^2 \right>=0.1124~m^2$ and $\sigma_{D(1)^2}^2=0.0026~m^4$ when 
$K_{\theta}=K_r=0.01m$ and $\left< D(1)^2 \right>=1.1443~m^2$ and $\sigma_{D(1)^2}^2=1.4681~m^4$ when $K_{\theta}=K_r=1m$.

\begin{figure}[htbp]
\begin{center}
\begin{tabular}{cc}
\includegraphics[width=.5 \columnwidth]{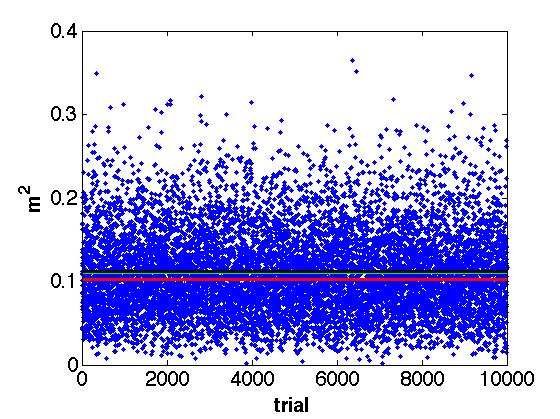}&
\includegraphics[width=.5 \columnwidth]{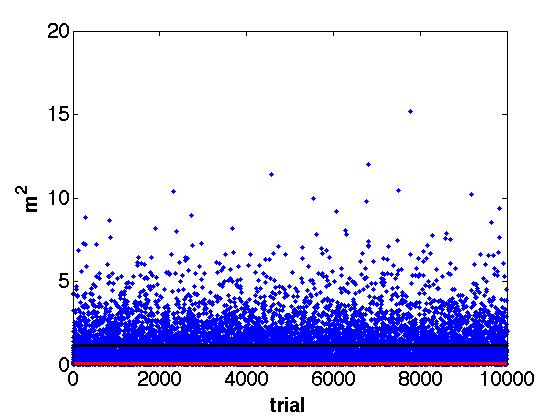}\\
\end{tabular}
\caption{Varying ratio between linear and angular speed. Values of $D(1)^2$ for $10^4$ trials together with the value of $D(1)^2$ for the trajectory without noise (red line), the value of $\left< D(1)^2 \right>$ obtained by averaging $D(1)^2$ on the trials (green line) and the value of $\left< D(1)^2 \right>$ obtained by our formulas (black line). $K_{\theta}=K_r=0.01m$ and $K_{\theta}=K_r=1m$ in the left and right image, respectively.}\label{FigStat}
\end{center}
\end{figure}

We ran $10^5$ trials and we computed from them the mean value and the variance. Figure \ref{FigStat} displays the values of $D(1)^2$ obtained in the first $10^4$ trials together with the value of $D(1)^2$ for the trajectory without noise (red line), the value of $\left< D(1)^2 \right>$ obtained by averaging $D(1)^2$ on the $10^4$ trials (green line) and the value of $\left< D(1)^2 \right>$ obtained by our formulas (black line). The image on the left refers to the case $K_{\theta}=K_r=0.01m$ and the image on the right to the case $K_{\theta}=K_r=1m$.

\noindent Table \ref{Table} reports the values of $\left< D(1)^2 \right>$ and $\sigma_{D(1)^2}^2$ obtained by running $10^3$, $10^4$ and $10^5$ trials. As in the case of constant ratio between linear and angular speed, we remark that, as the number of trials increases, the results converge to the values obtained by using our formulas.

\begin{table}
\begin{center}
\begin{tabular}{|c|c|c|c|}
  \hline
  $K_{\theta}=K_r~ (m)$  &  trials & $\left< D(1)^2 \right>~(m^2)$ & $\sigma_{D(1)^2}^2~(m^4)$ \\  
   \hline
   \hline
   $0.01$ & $10^3$ & $0.1139$ & $0.0022$\\
   \hline
   $0.01$ & $10^4$ & $0.1107$ & $0.0026$\\
   \hline
   $0.01$ & $10^5$ & $0.1123$ & $0.0026$\\
   \hline
   \hline
   $1$ & $10^3$ & $1.1094$ & $1.2534$\\
   \hline
    $1$ & $10^4$ & $1.1605$ & $1.5289$\\
   \hline
    $1$ & $10^5$ & $1.1494$ & $1.4708$\\
   \hline
   \hline
\end{tabular}
\end{center}
\caption{Varying ratio between linear and angular speed. Values of $\left< D(1)^2 \right>$ and $\sigma_{D(1)^2}^2$ obtained by running $10^3$, $10^4$ and $10^5$ trials.}
\label{Table}
\end{table}

\newpage

\section{Conclusion}\label{SectionConclusion}

In this paper we derived the statistics, up to any order, for $2D-$Brownian unicycle dynamics. The chosen kinematic constraint is modelled by the unicycle differential equation which is a very general constraint for a $2D$ motion. According to this model, the mobile robot can freely rotates but only one direction for the shift is allowed. This model is suitable to characterize the dynamics of many wheeled robots: namely, the ones that satisfy the unicycle dynamics. Additionally, the considered deterministic motion is very general. The expressions here provided for the statistics hold for any deterministic trajectory that satisfies the mentioned kinematic constraint. The expressions contain multiple integrals over the deterministic trajectory. 

We showed the power of the derived analytical results by investigating the diffusivity of the considered Brownian motion for constant and arbitrary time-dependent linear and angular speed.

We want to remark that this paper provides the analytical expressions for the statistics, up to any order, of a non-trivial Brownian motion, i.e. propelled by arbitrary force and torque. To the best of our knowledge, analytical solutions for any order moment in the case of arbitrary linear and angular speed have never been derived in the past.

\newpage


\appendix

\section{Computation of $\left<u(s)^p ~w(s)^q\right>$}\label{AppendixUV}

We have:

\begin{equation}\label{Equation<UV>}
\left<u(s)^p ~w(s)^q\right>=\lim_{N\rightarrow\infty} \sum_{j_1\cdots j_p k_1\cdots k_q} \left<e^{i(\theta_{j_1}+\cdots +\theta_{j_p} - \theta_{k_1}-\cdots -\theta_{k_q})}\right>
\end{equation}
\[
\left<(\delta s + \epsilon_{j_1})\cdots (\delta s + \epsilon_{j_p})(\delta s + \epsilon_{k_1})\cdots (\delta s + \epsilon_{k_q})\right>
\]

\noindent where each index goes from $1$ to $N$. Let us focus our attention on the quantity

\[
\left<(\delta s + \epsilon_{j_1})\cdots (\delta s + \epsilon_{j_p})(\delta s + \epsilon_{k_1})\cdots (\delta s + \epsilon_{k_q})\right>
\]
We can write this quantity as the sum of $p+q+1$ terms, i.e., $\delta s^{p+q}C_0 + \delta s^{p+q-1} C_1 + \delta s^{p+q-2} C_2 + \cdots + \delta s C_{p+q-1} + C_{p+q}$. We trivially have $C_0=1$. Concerning the remaining coefficients, we remark, first of all, that only the ones with even index are different from $0$. In other words, $C_{2n+1}=0$, $n=0,1,\cdots,\lfloor \frac{p+q}{2} \rfloor$
(where $ \lfloor x \rfloor $ is the largest integer not greater than $ x $). Let us compute $C_{2n}$. This coefficient is the sum of $ {{p+q}\choose{2n}} $ elements, each of them being the average of a product of $2n$ terms "$\epsilon$". On the other hand, since the exponential in (\ref{Equation<UV>}) is symmetric with respect to the change $j_a \leftrightarrow j_{a'}$, $a,a'=1,\cdots,p$ and with respect to the change $k_b \leftrightarrow k_{b'}$, $b,b'=1,\cdots,q$ but it is not symmetric with respect to the change $j_a \leftrightarrow k_b$, we need to separate the $ {{p+q}\choose{2n}} $ elements in several groups.
Specifically, let us denote by $l$ the number of $\epsilon$ with index of type $k$. The number of elements belonging to this group is $ {{p}\choose{2n-l}} {{q}\choose{l}} $. Note that we set ${{a}\choose{b}}=0$ when $b>a$. Note also that $\sum_{l=0}^{2n} {{p}\choose{2n-l}} {{q}\choose{l}} = {{p+q}\choose{2n}} $. We now remark that, because of the statistical properties of $\epsilon$, the average of the product of $2n$ terms $\epsilon$ is different from zero only when their indexes are equal two-by-two. Hence, we have to consider all the combinations of products of terms $\epsilon$, whose indexes are equal two-by-two and which differ for at least one pair of indexes. On the other hand, for a given element in the group characterized by a given $n$ and $l$, the effect in (\ref{Equation<UV>}) depends on the number of pairs which are {\it hetero} (i.e., with an index of type $j$ and one of type $k$). Let us denote by $m$ the number of pairs which are hetero. Note that $m$ has the same parity of $l$. Indeed, by definition we have $l$ indexes of type $k$. Additionally, we are using $m$ indexes of type $j$ and $m$ indexes of type $k$ to make $m$ hetero pairs. Hence, $l-m$ indexes of type $k$ remain and, with them, we have to make $\frac{l-m}{2}$ homo pairs of type $k$. Similarly, we have to make $\frac{2n-l-m}{2}$ homo pairs of type $j$. Hence, $\frac{l-m}{2}$ must be integer.

We compute the number of combinations of the products of $2n$ terms $\epsilon$, with $l$ indexes of type $k$, where the indexes are equal two-by-two, with $m$ pairs which are hetero and which differ for at least one pair. We obtain: ${{2n-l}\choose{m}} {{l}\choose{m}} m! (2n-l-m-1)!! (l-m-1)!!$. Following this, we can write (\ref{Equation<UV>}) as follows:

\begin{equation}\label{Equation<UV>Step1}
\left<u(s)^p ~w(s)^q\right>=\lim_{N\rightarrow\infty}  \sum_{\{j\}_p \{k\}_q} \sum_{n=0}^{\lfloor \frac{p+q}{2} \rfloor} \delta s^{p+q-2n}
\end{equation}
\[
\sum_{l=0}^{2n} {{p}\choose{2n-l}}{{q}\choose{l}} \sum_{m=0~odd/even}^{min(l,2n-l)} {{2n-l}\choose{m}} {{l}\choose{m}} m!
\]
\[
(2n-l-m-1)!! (l-m-1)!! \left<\epsilon_{j_1}^2\right>\delta_{j_1 j_2} \left<\epsilon_{j_3}^2\right>\delta_{j_3 j_4}\cdots
\]
\[
\left<\epsilon_{j_{2n-l-m-1}}^2\right>\delta_{j_{2n-l-m-1} j_{2n-l-m}} \left<\epsilon_{k_1}^2\right>\delta_{k_1 k_2} \left<\epsilon_{k_3}^2\right>\delta_{k_3 k_4}
\cdots
\]
\[
\left<\epsilon_{k_{l-m-1}}^2\right>\delta_{k_{l-m-1} k_{l-m}}
\left<\epsilon_{k_{l-m+1}}^2\right>\delta_{k_{l-m+1} j_{2n-l-m+1}}
\cdots \]
\[\left<\epsilon_{k_l}^2\right>\delta_{k_l j_{2n-l}}
\left<e^{i(\theta_{j_1}+\cdots +\theta_{j_p} - \theta_{k_1}-\cdots -\theta_{k_q})}\right>
\]

\noindent where, for the brevity sake, we denoted by $\sum_{\{j\}_p \{k\}_q}$ the sum $\sum_{j_1\cdots j_p k_1\cdots k_q=1}^N $. Each average $\left<\epsilon^2\right>$ provides $K_r \delta s$. Hence, all together, they provide $K_r^n \delta s^n$. Additionally, the number of Kronecker deltas is $n$. Hence, in the limit of $N\rightarrow\infty$ for each value of $n$ we get a multiple integral of dimension $p+q-n$. Note that, when the indexes are equal h-by-h ($h>2$), the result in the limit $N\rightarrow\infty$ vanishes since, the power of $\delta s$, is larger than the number of sums. By a direct computation in (\ref{Equation<UV>Step1}) we obtain:

\begin{equation}\label{Equation<UV>Step2}
\left<u(s)^p ~w(s)^q\right>=\sum_{n=0}^{\lfloor \frac{p+q}{2} \rfloor} K_r^n \sum_{l=0}^{2n} {{p}\choose{2n-l}} {{q}\choose{l}}\sum_{m=0~odd/even}^{min(l,2n-l)}
\end{equation}
\[
{{2n-l}\choose{m}} {{l}\choose{m}} m! (2n-l-m-1)!! (l-m-1)!! ~s^m \lim_{N\rightarrow\infty} \]
\[\sum_{\{j^s\}_{\sigma} \{j^d\}_{\rho} \{k^s\}_{\chi} \{k^d\}_{\eta} } \delta s^{\beta} \left<exp\left\{ i[\theta_{j^s_1}+\cdots +\theta_{j^s_{\sigma}} +2 (\theta_{j^d_1}+ \right. \right. \]
\[
 \left. \left.  +\cdots+ \theta_{j^d_{\rho}}) - (\theta_{k^s_1}+\cdots +\theta_{k^s_{\chi}}) -2 (\theta_{k^d_1}+\cdots +\theta_{k^d_{\eta}})]\right\}\right>
\]

\noindent where $\rho \equiv\frac{2n-l-m}{2}$, $\sigma \equiv p-(2n-l)$, $\eta \equiv \frac{l-m}{2}$, $\chi \equiv q-l$ and $\beta \equiv \rho+\sigma+\eta+\chi=p+q-n-m$. We must compute the average of the exponential in (\ref{Equation<UV>Step2}) and then we compute the limit $N\rightarrow\infty$. We remark that the various $\theta$ in the exponential contain random quantities (i.e., the $\delta\theta$ at different time steps). In order to proceed we have to separate all the random quantities which are independent. We start this separation by redefining the indexes in the sum $\sum_{\{j^s\}_{\sigma} \{j^d\}_{\rho} \{k^s\}_{\chi} \{k^d\}_{\eta}}$. Specifically, we consider the new indexes $\overline{j}^s \overline{j}^d \overline{k}^s \overline{k}^d$ which differ from $j^s j^d k^s k^d$ since they are ordered (in increasing order). For instance, $\overline{j}^s_1 < \overline{j}^s_2 < \cdots < \overline{j}^s_{\sigma}$. Hence, the last sum in (\ref{Equation<UV>Step2}) can be replaced with
$\sum_{\{j^s\}_{\sigma} \{j^d\}_{\rho} \{k^s\}_{\chi} \{k^d\}_{\eta}} \rightarrow \rho! \sigma! \eta! \chi! \sum_{\{\overline{j}^s\}_{\sigma} \{\overline{j}^d\}_{\rho} \{\overline{k}^s\}_{\chi} \{\overline{k}^d\}_{\eta} }$. The four types of indexes are not ordered among them. Hence, the sum includes all the possible combinations which maintain the order only restricted to a single index type. For instance, a possible combination is: $\overline{k}^d_1 <\overline{k}^d_2 <\overline{j}^s_1 <\overline{k}^s_1 <\overline{j}^s_2 <\overline{k}^d_3 <\overline{j}^d_1 <\overline{j}^d_2 <\cdots$. We introduce the $\beta$ ordered indexes $\overline{w}_1<\overline{w}_2<\cdots<\overline{w}_{\beta}$. Additionally, let us denote with $\Delta_a^b \equiv \sum_{c=a+1}^b \delta \theta_c$ and $\overline{\Delta}_a^b \equiv \sum_{c=a+1}^b \overline{\delta \theta}_c = \overline{\theta}(b\delta s) - \overline{\theta}(a\delta s)$, where $\overline{\theta}(s)$ is the deterministic trajectory. Finally, let us define $\alpha \equiv 2\rho+\sigma - \chi -2\eta$. The sum in the exponential in (\ref{Equation<UV>Step2}) contains $\alpha (\Delta_0^{\overline{w}_1}+\overline{\Delta}_0^{\overline{w}_1})$. Then, depending on the combination of the indexes $\overline{j}^s \overline{j}^d \overline{k}^s \overline{k}^d$, we have a different result for the $(\Delta_{\overline{w}_1}^{\overline{w}_2}+\overline{\Delta}_{\overline{w}_1}^{\overline{w}_2})$. Specifically, if $\overline{w}_1={\overline{j}^d_1}$ the sum in the exponential contains $(\alpha -2)(\Delta_{\overline{w}_1}^{\overline{w}_2}+\overline{\Delta}_{\overline{w}_1}^{\overline{w}_2})$. If $\overline{w}_1={\overline{j}^s_1}$ it contains $(\alpha -1)(\Delta_{\overline{w}_1}^{\overline{w}_2}+\overline{\Delta}_{\overline{w}_1}^{\overline{w}_2})$.  If $\overline{w}_1={\overline{k}^s_1}$ it contains $(\alpha +1)(\Delta_{\overline{w}_1}^{\overline{w}_2}+\overline{\Delta}_{\overline{w}_1}^{\overline{w}_2})$.  If $\overline{w}_1={\overline{k}^d_1}$ it contains $(\alpha +2)(\Delta_{\overline{w}_1}^{\overline{w}_2}+\overline{\Delta}_{\overline{w}_1}^{\overline{w}_2})$. We introduce the vector $\mbox{\boldmath $c$}\equiv [c_1, \cdots,c_{\beta}]^T$ whose entries are $-2$, $-1$, $1$ and $2$. Specifically, it contains $\eta$ entries equal to $2$, $\chi$ equal to $1$, $\sigma$ equal to $-1$ and $\rho$ equal to $-2$. It is easy to realize that we have ${{\beta}\choose{\chi}} {{\beta-\chi}\choose{\sigma}} {{\rho+\eta}\choose{\eta}}$ vectors $\mbox{\boldmath $c$}$. Finally, we define $\Phi_b \equiv \sum_{a=1}^b c_a$. Note that $\Phi_{\beta}=-\alpha$.
According to this, the last sum in (\ref{Equation<UV>Step2}) can be written as follows:

\[
e^{i\alpha\theta_0}\rho! \sigma! \eta! \chi! \sum_{\mbox{\boldmath $c$}}\sum_{\left\{\overline{w}\right\}_{\beta}}\delta s^{\beta} \cdot
 \]
 \[
 \cdot \left<exp\left\{i \sum_{b=0}^{\beta-1}\left[ (\alpha + \Phi_b) (\Delta_{\overline{w}_b}^{\overline{w}_{b+1}}+\overline{\Delta}_{\overline{w}_b}^{\overline{w}_{b+1}})\right]\right\}\right>
 \]

 \noindent Hence, we have:

\begin{equation}\label{Equation<UV>Step3}
\left<u(s)^p ~w(s)^q\right>=\sum_{n=0}^{\lfloor \frac{p+q}{2} \rfloor} K_r^n \sum_{l=0}^{2n} {{p}\choose{2n-l}} {{q}\choose{l}} \cdot
\end{equation}
\[
\cdot \sum_{m=0~odd/even}^{min(l,2n-l)} {{2n-l}\choose{m}} {{l}\choose{m}} m! (2n-l-m-1)!! (l-m-1)!! \cdot
\]
\[
\cdot s^m  e^{i\alpha\theta_0}\rho! \sigma! \eta! \chi! \sum_{\mbox{\boldmath $c$}}\lim_{N\rightarrow\infty} \sum_{\left\{\overline{w}\right\}_{\beta}}\delta s^{\beta} \cdot
 \]
 \[ \cdot \left<exp\left\{i \sum_{b=0}^{\beta-1}\left[ (\alpha + \Phi_b) (\Delta_{\overline{w}_b}^{\overline{w}_{b+1}}+\overline{\Delta}_{\overline{w}_b}^{\overline{w}_{b+1}})\right]\right\}\right>
\]

\noindent Now we can compute the average since we were able to separate all the independent quantities. We have:

\begin{equation}\label{EquationAppendixIntermediate}
\left<exp\left\{i \sum_{b=0}^{\beta-1}\left[ (\alpha + \Phi_b) (\Delta_{\overline{w}_b}^{\overline{w}_{b+1}}+\overline{\Delta}_{\overline{w}_b}^{\overline{w}_{b+1}})\right]\right\}\right>=
\end{equation}
\[
=exp\left\{i \sum_{b=0}^{\beta-1}\left[ (\alpha + \Phi_b) \overline{\Delta}_{\overline{w}_b}^{\overline{w}_{b+1}}\right]\right\} \cdot \]
\[ \cdot \prod_{b=0}^{\beta-1}\left<exp\left\{i \left[ (\alpha + \Phi_b) \Delta_{\overline{w}_b}^{\overline{w}_{b+1}}\right]\right\}\right>=
\]
\[
=exp\left\{i \sum_{b=0}^{\beta-1}\left[ (\alpha + \Phi_b) \overline{\Delta}_{\overline{w}_b}^{\overline{w}_{b+1}}\right]\right\} \cdot \]
\[ \cdot \prod_{b=0}^{\beta-1} exp\left\{-\frac{(\alpha + \Phi_b)^2 (\overline{w}_{b+1}-\overline{w}_b)K_{\theta}\delta s}{2}  \right\}=
\]
\[
e^{\sum_{b=0}^{\beta-1}\left[ i (\alpha + \Phi_b) \overline{\Delta}_{\overline{w}_b}^{\overline{w}_{b+1}}-\frac{(\alpha + \Phi_b)^2 (\overline{w}_{b+1}-\overline{w}_b)K_{\theta}\delta s}{2}  \right] }
\]

\noindent We used the equality ($\eta=\mathcal{N}\left(0, 1\right)$):

\begin{equation}\label{EquationAppendixRule1}
\left<e^{A\eta}\right>=e^{\frac{A^2}{2}}
\end{equation}

\noindent with $A=i  (\alpha + \Phi_b)\sigma_{b+1}$ and $\sigma^2_{b+1}=(\overline{w}_{b+1}-\overline{w}_b)K_{\theta}\delta s$. By substituting equation (\ref{EquationAppendixIntermediate}) in (\ref{Equation<UV>Step3}) and by taking the limit ($N\rightarrow\infty$), we finally obtain:

\begin{equation}\label{Equation<UV>ComputedAppendix}
\left<u(s)^p ~w(s)^q\right>=\sum_{n=0}^{\lfloor \frac{p+q}{2} \rfloor} K_r^n \sum_{l=0}^{2n} {{p}\choose{2n-l}} {{q}\choose{l}} \cdot
\end{equation}
\[
\cdot \sum_{m=0~odd/even}^{min(l,2n-l)} {{2n-l}\choose{m}} {{l}\choose{m}}m! (2n-l-m-1)!! (l-m-1)!! \cdot
\]
\[\cdot s^m e^{i\alpha\theta_0}\rho! \sigma! \eta! \chi! \sum_{\mbox{\boldmath $c$}} \int_0^s ds_1 \int_{s_1}^s ds_2 \cdots \int_{s_{\beta-1}}^s ds_{\beta} \]
\[
e^{ \sum_{b=0}^{\beta-1}\left[ i(\alpha + \Phi_b) [\overline{\theta}(s_{b+1})-\overline{\theta}(s_b)]-\frac{(\alpha + \Phi_b)^2 (s_{b+1}-s_b)K_{\theta}}{2}\right]}
\]

\noindent which coincides with (\ref{Equation<UV>Computed}).

\section{Computation of $\left<u(s)^p ~w(s)^q~\tilde{\theta}^r\right>$}\label{AppendixUVTheta}

This computation follows the same initial steps carried out in appendix \ref{AppendixUV}. We obtain an expression equal to the one given in (\ref{Equation<UV>Step3}) but the mean value at the end must be replaced with $\left<exp\left\{i \sum_{b=0}^{\beta-1}\left[ (\alpha + \Phi_b) (\Delta_{\overline{w}_b}^{\overline{w}_{b+1}}+\overline{\Delta}_{\overline{w}_b}^{\overline{w}_{b+1}})\right]\right\} \cdot \tilde{\theta}^r\right>$.
The deterministic part can be factorized out the mean value. We need to calculate:

\[
\left<exp\left\{i \sum_{b=0}^{\beta-1}\left[ (\alpha + \Phi_b) \Delta_{\overline{w}_b}^{\overline{w}_{b+1}}\right]\right\} \cdot \tilde{\theta}^r\right>=
\]
\[=\left< \tilde{\theta}^r\prod_{b=0}^{\beta-1}\left\{exp\left[i (\alpha + \Phi_b) \Delta_{\overline{w}_b}^{\overline{w}_{b+1}}\right]\right\} \right>
\]

\noindent By using the multinomial theorem we can write:

\[
\tilde{\theta}^r=\left(\sum_{b=0}^{\beta}\Delta_{\overline{w}_b}^{\overline{w}_{b+1}}\right)^r=\sum_{\mbox{\boldmath $\gamma$}}r!\cdot \prod_{b=0}^{\beta}\frac{\left(\Delta_{\overline{w}_b}^{\overline{w}_{b+1}}\right)^{\gamma_{b+1}}}{\gamma_{b+1}!}
\]

\noindent where $\overline{w}_{\beta+1}\equiv N$ and the sum over $\mbox{\boldmath $\gamma$}$ is the sum over all the vectors $\mbox{\boldmath $\gamma$}=[\gamma_1,\cdots,\gamma_{\beta+1}]$, where $\gamma_b$ ($b=1,\cdots,\beta+1$) are positive integers satisfying the constraint: $\sum_{b=1}^{\beta+1}\gamma_b=r$. Hence, we obtain

\begin{equation}\label{EquationAppendixStepA}
\left<\tilde{\theta}^r\prod_{b=0}^{\beta-1}\left\{e^{i (\alpha + \Phi_b) \Delta_{\overline{w}_b}^{\overline{w}_{b+1}}}\right\} \right>=
\end{equation}
\[\sum_{\mbox{\boldmath $\gamma$}}r!
\left<\frac{\left(\Delta_{\overline{w}_{\beta}}^{\overline{w}_{\beta+1}}\right)^{\gamma_{\beta+1}}}{\gamma_{\beta+1}!}\prod_{b=0}^{\beta-1}e^{i (\alpha + \Phi_b) \Delta_{\overline{w}_b}^{\overline{w}_{b+1}}} \frac{\left(\Delta_{\overline{w}_b}^{\overline{w}_{b+1}}\right)^{\gamma_{b+1}}}{\gamma_{b+1}!}\right>=
\]
\[=\sum_{\mbox{\boldmath $\gamma$}}\frac{r!}{\prod_{b=0}^{\beta}\gamma_{b+1}!}
\left<\left(\Delta_{\overline{w}_{\beta}}^{\overline{w}_{\beta+1}}\right)^{\gamma_{\beta+1}}\right> \cdot \]
\[\cdot \prod_{b=0}^{\beta-1}\left<e^{i (\alpha + \Phi_b) \Delta_{\overline{w}_b}^{\overline{w}_{b+1}}} \left(\Delta_{\overline{w}_b}^{\overline{w}_{b+1}}\right)^{\gamma_{b+1}}\right>
\]

\noindent We use the following two standard results for a normal distribution:

\begin{equation}\label{EquationAppendixRule2}
\left<\left(\Delta_{\overline{w}_{\beta}}^{\overline{w}_{\beta+1}}\right)^{\gamma_{\beta+1}}\right>=\left[
\begin{array}{ccc}
0&\gamma_{\beta+1}&odd\\
\sigma_{\beta +1}^{\gamma_{\beta+1}} (\gamma_{\beta+1}-1)!!&\gamma_{\beta+1}&even\\
\end{array}
\right.
\end{equation}

\noindent and

\begin{equation}\label{EquationAppendixRule3}
\left<exp\left[i (\alpha + \Phi_b) \Delta_{\overline{w}_b}^{\overline{w}_{b+1}}\right] \left(\Delta_{\overline{w}_b}^{\overline{w}_{b+1}}\right)^{\gamma_{b+1}}\right>=
\end{equation}
\[
=\sigma_{b+1}^{\gamma_{b+1}}exp\left[-\frac{(\alpha + \Phi_b)^2 \sigma_{b+1}^2}{2} \right] \cdot \]
\[\cdot \sum_{a=0}^{\lfloor\frac{\gamma_{b+1}}{2}\rfloor}\frac{\gamma_{b+1}!\left(i (\alpha + \Phi_b) \sigma_{b+1}\right)^{\gamma_{b+1}-2a}}{a!(\gamma_{b+1}-2a)!2^a}
\]

\noindent with $\sigma_{b+1}^2 \equiv K_{\theta} \delta s (\overline{w}_{b+1}-\overline{w}_b)$. Equation (\ref{EquationAppendixRule3}) is obtained starting from (\ref{EquationAppendixRule1}) and by differentiating $\gamma_{b+1}$ times with respect to $A$. Hence, the expression in (\ref{EquationAppendixStepA}), becomes:

\begin{equation}\label{EquationAppendixStepB}
\left<\tilde{\theta}^r\prod_{b=0}^{\beta-1}\left\{exp\left[i (\alpha + \Phi_b) \Delta_{\overline{w}_b}^{\overline{w}_{b+1}}\right]\right\} \right>=
\end{equation}
\[
=\sum_{\mbox{\boldmath $\gamma$}}\frac{r!\sigma_{\beta +1}^{\gamma_{\beta+1}} (\gamma_{\beta+1}-1)!!}{\prod_{b=0}^{\beta}\gamma_{b+1}!}
\prod_{b=0}^{\beta-1}
\left\{\sigma_{b+1}^{\gamma_{b+1}}e^{-\frac{(\alpha + \Phi_b)^2 \sigma_{b+1}^2}{2}}
\cdot \right.\]
\[ \left. \cdot \sum_{a=0}^{\lfloor\frac{\gamma_{b+1}}{2}\rfloor}\frac{\gamma_{b+1}!\left(i (\alpha + \Phi_b) \sigma_{b+1}\right)^{\gamma_{b+1}-2a}}{a!(\gamma_{b+1}-2a)!2^a}\right\}
\]

\noindent where now the sum over $\mbox{\boldmath $\gamma$}$ only includes the vectors $\mbox{\boldmath $\gamma$}$ whose last entry is even. By taking the limit $N\rightarrow\infty$, this expression becomes:

\[
\sum_{\mbox{\boldmath $\gamma$}}\frac{r! (\gamma_{\beta+1}-1)!!}{\prod_{b=0}^{\beta}\gamma_{b+1}!}(s-s_{\beta})^{\frac{\gamma_{\beta+1}}{2}}K_{\theta}^{\frac{r}{2}} \cdot
\]
\[
\cdot \prod_{b=0}^{\beta-1}
\left\{ (s_{b+1}-s_b)^{\frac{\gamma_{b+1}}{2}}e^{-\frac{(\alpha + \Phi_b)^2 K_{\theta} (s_{b+1}-s_b)}{2}}
\cdot \right.\]
\[ \left. \cdot \sum_{a=0}^{\lfloor\frac{\gamma_{b+1}}{2}\rfloor}\frac{\gamma_{b+1}!\left(i (\alpha + \Phi_b) \sqrt{K_{\theta} (s_{b+1}-s_b)}\right)^{\gamma_{b+1}-2a}}{a!(\gamma_{b+1}-2a)!2^a}\right\}
\]

\noindent and we finally obtain:

\begin{equation}\label{Equation<UVT>ComputedAppendix}
\left<u(s)^p ~w(s)^q~\tilde{\theta}(s)^r\right>=\sum_{n=0}^{\lfloor \frac{p+q}{2} \rfloor} K_r^n \sum_{l=0}^{2n} {{p}\choose{2n-l}} {{q}\choose{l}} \cdot
\end{equation}
\[
\cdot \sum_{m=0~odd/even}^{min(l,2n-l)} {{2n-l} \choose{m}}{{l}\choose{m}}m! (2n-l-m-1)!! (l-m-1)!!  \]
\[
\cdot s^m e^{i\alpha\theta_0}\rho! \sigma! \eta! \chi! \sum_{\mbox{\boldmath $c$}} \sum_{\mbox{\boldmath $\gamma$}}\frac{r! (\gamma_{\beta+1}-1)!! K_{\theta}^{\frac{r}{2}}}{\prod_{b=0}^{\beta}\gamma_{b+1}!}\int_0^s ds_1 \int_{s_1}^s ds_2 \cdot \]
\[
\cdots \int_{s_{\beta-1}}^s ds_{\beta}
(s-s_{\beta})^{\frac{\gamma_{\beta+1}}{2}}
e^{ i\sum_{b=0}^{\beta-1}(\alpha + \Phi_b) [\overline{\theta}(s_{b+1})-\overline{\theta}(s_b)]}
\]
\[
\prod_{b=0}^{\beta-1}
\left\{ (s_{b+1}-s_b)^{\frac{\gamma_{b+1}}{2}}e^{-\frac{(\alpha + \Phi_b)^2 K_{\theta} (s_{b+1}-s_b)}{2}} \cdot \right.\]
\[ \left.\sum_{a=0}^{\lfloor\frac{\gamma_{b+1}}{2}\rfloor}\frac{\gamma_{b+1}!\left(i (\alpha + \Phi_b) \sqrt{K_{\theta} (s_{b+1}-s_b)}\right)^{\gamma_{b+1}-2a}}{a!(\gamma_{b+1}-2a)!2^a}\right\}
\]

\noindent which coincides with (\ref{Equation<UVT>Computed}).

\section{Computation of $\left<D(s)^4\right>=\left<u(s)^2 ~w(s)^2\right>$}\label{AppendixD4}

We use equation (\ref{Equation<UV>Computed}) with $p=q=2$ to obtain the analytical expression of $\left< D(s)^4\right>= \left< u(s)^2w(s)^2\right>$. The first sum on $n$ includes the three terms $n=0,1,2$. When $n=0$, the second sum on $l$ and the third sum on $m$ only include one term, i.e., $l=m=0$. All the factors before the sum on 
$\mbox{\boldmath $c$}$ are $1$ with the exception of $(p-2n+l)!=2$ and $(q-l)!=2$. Additionally, $\beta=4$ and the sum on $\mbox{\boldmath $c$}$ consists of six vectors, i.e., $[-1, -1, 1, 1]$, $[-1, 1, -1, 1]$, $[-1, 1, 1, -1]$, $[1, -1, -1, 1]$, $[1, -1, 1, -1]$, $[1, 1, -1, -1]$. Hence we obtain the following term:

\[
C_{000}=8\int_0^s ds_1 \int_{s_1}^s ds_2 \int_{s_2}^s ds_3 \int_{s_3}^s ds_4\left\{ e^{-\frac{K_{\theta}(-s_1-3s_2+3s_3+s_4)}{2}} \cos\left(-\overline{\theta}_1-\overline{\theta}_2+\overline{\theta}_3+\overline{\theta}_4\right) +\right.
\]
\[
\left.   + e^{-\frac{K_{\theta}(-s_1+s_2-s_3+s_4)}{2}} \left[  \cos\left(-\overline{\theta}_1+\overline{\theta}_2-\overline{\theta}_3+\overline{\theta}_4\right) +\cos\left(-\overline{\theta}_1+\overline{\theta}_2+\overline{\theta}_3-\overline{\theta}_4\right)\right] \right\}
\]

\noindent where $\overline{\theta}_j\equiv\overline{\theta}(s_j)$, $j=1,2,3,4$.

Let us consider now the terms with $n=1$. In this case, $l=0,1,2$. For each value of $l$ we have in general several values of $m$, which must have the same parity of $l$. Hence, when $l=1$, we only have $m=1$. When $l=2$, we can have $m=0,2$. On the other hand, when $l=m=2$, the factor ${{2n-l}\choose{m}}={{0}\choose{2}}$ vanishes. By an explicit computation, it is possible to see that $C_{120}$ is the conjugate of $C_{100}$. Hence, their sum is real and it is:

\[
C_{100}+C_{120}=4K_r\int_0^s ds_1 \int_{s_1}^s ds_2 \int_{s_2}^s ds_3 \left\{ e^{-\frac{K_{\theta}(-4s_1+3s_2+s_3)}{2}} \cos\left(2\overline{\theta}_1-\overline{\theta}_2-\overline{\theta}_3\right) +\right.
\]
\[
\left.   + e^{-\frac{K_{\theta}(-s_1+s_3)}{2}} \cos\left(-\overline{\theta}_1+2\overline{\theta}_2-\overline{\theta}_3\right) +e^{-\frac{K_{\theta}(-s_1-3s_2+4s_3)}{2}} \cos\left(-\overline{\theta}_1-\overline{\theta}_2+2\overline{\theta}_3\right) \right\}
\]

\noindent The last term with $n=1$ is the one with $l=m=1$. By a direct computation we obtain:

\[
C_{111}=8K_r s\int_0^s ds_1 \int_{s_1}^s ds_2 e^{-\frac{K_{\theta}(-s_1+s_2)}{2}} \cos\left(-\overline{\theta}_1+\overline{\theta}_2\right) 
\]

\noindent Finally, when $n=2$ the sum on $l$ includes the five values: $l=0,1,2,3,4$. On the other hand, only the term $l=2$ does not vanish since the product ${{p}\choose{2n-l}} {{q}\choose{l}}$ is zero in all the other cases. When $l=2$ we have two possible values for $m$, i.e., $m=0,2$. We obtain the following two contributions:

\[
C_{220}=2K_r^2\int_0^s ds_1 \int_{s_1}^s ds_2 e^{-2K_{\theta}(-s_1+s_2)} \cos\left(-2\overline{\theta}_1+2\overline{\theta}_2\right) 
\]

\[
C_{222}=2K_r^2s^2 
\]

\end{document}